\documentclass[a4paper,fleqn]{cas-dc}

\usepackage[numbers]{natbib}

\usepackage[ruled,vlined]{algorithm2e}
\usepackage{acronym}
\usepackage{mathtools}
\usepackage{float}

\acrodef{WSI}[WSI]{Whole-Slide Image}
\acrodef{DRE}[DRE]{Digital Rectal Exam}
\acrodef{PSA}[PSA]{Prostate-Specific Antigen}
\acrodef{WHO}[WHO]{World Health Organization}
\acrodef{AI}[AI]{Artificial Intelligence}
\acrodef{CNN}[CNN]{Convolutional Neural Network}
\acrodef{CAD}[CAD]{Computer-Aided Diagnosis}
\acrodef{DL}[DL]{Deep Learning}
\acrodef{ML}[ML]{Machine Learning}
\acrodef{NN}[NN]{Neural Network}
\acrodef{H&E}[H\&E]{Hematoxylin and Eosin}
\acrodef{W&D}[W\&D]{Wide \& Deep}
\acrodef{ANN}[ANN]{Artificial Neural Network}
\acrodef{RNN}[RNN]{Recurrent Neural Network}
\acrodef{SVM}[SVM]{Support Vector Machine}
\acrodef{RF}[RF]{Random Forest}
\acrodef{KNN}[KNN]{k-Nearest Neighbors}
\acrodef{AUC}[AUC]{Area Under Curve}
\acrodef{ROC}[ROC]{Receiver Operating Characteristic}
\acrodef{PCa}[PCa]{prostate cancer}

\def\tsc#1{\csdef{#1}{\textsc{\lowercase{#1}}\xspace}}
\tsc{WGM}
\tsc{QE}
\tsc{EP}
\tsc{PMS}
\tsc{BEC}
\tsc{DE}

\begin{document}
\let\WriteBookmarks\relax
\def\floatpagepagefraction{1}
\def\textpagefraction{.001}
\shorttitle{Wide \& Deep neural network model for patch aggregation in CNN-based prostate cancer detection systems}
\shortauthors{L. Duran-Lopez et~al.}

\title [mode = title]{Wide \& Deep neural network model for patch aggregation in CNN-based prostate cancer detection systems}                      



\address[1]{Robotics and Tech. of Computers Lab., Universidad de Sevilla, 41012 Seville, Spain}
\address[2]{Escuela Técnica Superior de Ingeniería Informática (ETSII), Universidad de Sevilla, 41012 Seville, Spain}
\address[3]{Escuela Politécnica Superior (EPS), Universidad de Sevilla, 41011 Seville, Spain}
\address[4]{Smart Computer Systems Research and Engineering Lab (SCORE), Research Institute of Computer Engineering (I3US), Universidad de Sevilla, 41012 Seville, Spain}

\author[1,2,3,4]{L. Duran-Lopez}[orcid=0000-0002-5849-8003]
\ead{lduran@atc.us.es}
\cormark[1]
\credit{Conceptualization, Data curation, Formal Analysis, Investigation, Methodology, Software, Validation, Visualization, Writing - original draft, Writing - review \& editing}

\author[1,2,3,4]{Juan P. Dominguez-Morales}[orcid=0000-0002-5474-107X]
\credit{Conceptualization, Investigation, Methodology, Project administration, Supervision, Validation, Visualization, Writing - original draft, Writing - review \& editing}

\author[1,2,3,4]{D. Gutierrez-Galan}[orcid=0000-0003-3100-0604]
\credit{Data curation, Validation, Visualization, Writing - review \& editing}

\author[1,2,3,4]{A. Rios-Navarro}[orcid=0000-0003-4163-8484]
\credit{Formal Analysis, Validation, Visualization, Writing - review \& editing}

\author[1,2,3,4]{A. Jimenez-Fernandez}[orcid=0000-0003-3061-5922]
\credit{Methodology, Validation, Visualization, Writing - review \& editing}

\author[1,2,3,4]{S. Vicente-Diaz}[orcid=0000-0001-9466-485X]
\credit{Funding acquisition, Project administration, Resources, Writing - review \& editing}

\author[1,2,3,4]{A. Linares-Barranco}[orcid=0000-0002-6056-740X]
\credit{Funding acquisition, Project administration, Resources, Writing - review \& editing}

\cortext[cor1]{Corresponding author}


\begin{abstract}
Prostate cancer (PCa) is one of the most commonly diagnosed cancer and one of the leading causes of death among men, with almost 1.41 million new cases and around 375,000 deaths in 2020. Artificial Intelligence algorithms have had a huge impact in medical image analysis, including digital histopathology, where Convolutional Neural Networks (CNNs) are used to provide a fast and accurate diagnosis, supporting experts in this task. To perform an automatic diagnosis, prostate tissue samples are first digitized into gigapixel-resolution whole-slide images. Due to the size of these images, neural networks cannot use them as input and, therefore, small subimages called patches are extracted and predicted, obtaining a patch-level classification. In this work, a novel patch aggregation method based on a custom Wide \& Deep neural network model is presented, which performs a slide-level classification using the patch-level classes obtained from a CNN. The malignant tissue ratio, a 10-bin malignant probability histogram, the least squares regression line of the histogram, and the number of malignant connected components are used by the proposed model to perform the classification. An accuracy of 94.24\% and a sensitivity of 98.87\% were achieved, proving that the proposed system could aid pathologists by speeding up the screening process and, thus, contribute to the fight against PCa.
\end{abstract}


\maketitle

\section{Introduction}

According to GLOBOCAN, \ac{PCa} is the second most frequently diagnosed cancer and the fifth leading cause of cancer death in men, with more than 1.41 million cases in 2020 and around 375,000 deaths worldwide \cite{sung2021global}. It is estimated that \ac{PCa} cases will increase with around 1,000,000 new cases in 2040, according to the \ac{WHO} \cite{rawla2019epidemiology}. 

Generally, the first step to diagnose \ac{PCa} consists in a \ac{DRE}, which is the primary test for the initial clinical assessment of the prostate. If an abnormal result for \ac{DRE} is found, a \ac{PSA} analysis is performed as a screening method for the investigation of a tumor. Then, in case of a positive \ac{PSA}, a trans-rectal ultrasound-guided biopsy is considered, which is the most certainly test to confirm or exclude the presence of cancer \cite{borley2009prostate}. With this technique, prostate samples are obtained, which are processed in a laboratory and scanned, resulting on gigapixel-resolution images called \acp{WSI}. These images are analyzed by pathologists to provide a final diagnosis with the corresponding cancer treatment.


The use of \ac{AI} in image analysis has had a huge impact in recent years \cite{hamet2017artificial, ahuja2019impact}, mainly due to the computational advances and the accessibility of its algorithms to researchers. Its application in the biomedical field has been expanded considerably, particularly, the use of \ac{DL}, which has become one of the most popular \ac{AI} techniques for image recognition in the last years \cite{shen2017deep}. These algorithms could play an important role as screening methods in order to report a second opinion and assist doctors in a specific image analysis tasks \cite{litjens2017survey, fourcade2019deep}. Particularly, this approach has recently been widely used in digital histopathology, where \acp{CNN} and other different \ac{DL} mechanisms are trained to analyze and detect malignant tissue in \acp{WSI}. Since \acp{CNN} cannot use an entire \ac{WSI} as input due to their high resolution, which would require a huge memory and processing capacity, a common approach to this problem is to extract smaller subimages from them, called patches. Therefore, the \ac{CNN} is able to analyze the \acp{WSI} at patch level and then report the classification results obtained.


Previous works, such as \cite{strom2020artificial, campanella2019clinical, litjens2016deep, li2018path, bulten2020automated}, have followed this patch-level classification strategy in order to develop \ac{DL}-based \ac{CAD} systems for \ac{PCa} detection in digitized histopathological images, reporting accurate results with different metrics and datasets. Among them, to the best of the authors' knowledge, PROMETEO \cite{duran2020prometeo} achieved the fastest and least complex model \cite{benchmark_prometeo} while also obtaining state-of-the-art results, leading to the most-plausible edge-computing solution for \ac{PCa} detection. This was achieved by means of a 9-layer custom \ac{CNN} trained and validated with a set of patches after applying different processing steps, including patch filtering, stain normalization and data augmentation. This allowed achieving 99.98\% accuracy, 99.98\% F1 score and 0.999 \ac{AUC} on a separate test set.


Since the results obtained when using \acp{CNN} are reported at patch level, different techniques have been proposed in the literature in order to combine them and generate a slide-level classification result, which could be of great importance for developing a fast \ac{PCa} screening system. This technique is known as patch aggregation. Among the different studies that can be found in the literature, some performed different patch aggregation techniques based on \acp{RNN} \cite{campanella2019clinical}, \acp{RF} \cite{campanella2019clinical} an other \ac{ML} or statistical alternatives \cite{strom2020artificial, bulten2020automated}, achieving accurate solutions and leading to precise screening methods that could help pathologists in their task.  

In this work, a custom novel Wide \& Deep (W\&D) model for aggregating the patch-level classification results obtained from the PROMETEO \ac{CAD} system into a global slide-level class is presented. This approach allows providing a fast screening method for \ac{PCa} detection at \ac{WSI} level, while also benefiting from the spatial resolution obtained at patch level. The promising results obtained, which have also been compared to other state-of-the-art \ac{ML}-based approaches, show that the proposed solution could aid pathologists when analyzing histopathological images, discriminating between positive and negative \ac{PCa} samples while fastening up the whole process.

The main contributions of this work include the following:

\begin{itemize}
  \item A set of algorithms to automatically extract relevant features from the output of a patch-level \ac{DL}-based \ac{PCa} detection system.
  \item A 5-layer custom W\&D model, trained and validated from scratch, which extracts independent features from the inputs and combine them to achieve a slide-level \ac{PCa} screening method with high sensitivity.
  \item A comparative study between different widely-known \ac{ML} algorithms for the patch-aggregation task on the same dataset, which shows that the proposed method achieves the highest sensitivity.
\end{itemize}

The rest of the paper is structured as follows: in section~\ref{sec:materials_and_methods}, the materials and methods are presented, focusing on the dataset that was used for this work (\ref{subsec:dataset}), along with the neural network model (\ref{subsec:wide_and_deep}), and the details on the training and validation steps of the model with the aforementioned dataset (\ref{subsec:training_and_validation}). Then, section~\ref{sec:results_and_discussion} presents the results obtained with the proposed model using different evaluation metrics. A comparison with other state-of-the-art \ac{ML} techniques is also performed in the  same section. Finally, the conclusions of this work are presented in section~\ref{sec:conclusions}.

\section{Materials and Methods}
\label{sec:materials_and_methods}
\subsection{Dataset}
\label{subsec:dataset}

\begin{figure*}[pos=htbp,width=\textwidth,align=\centering]
\centering\includegraphics[width=1\textwidth]{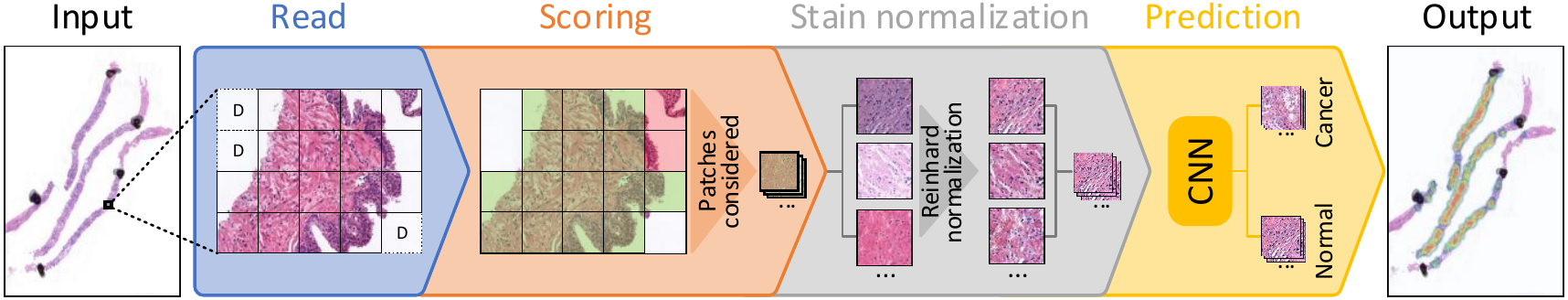}
\caption{\textbf{Block diagram detailing each of the steps considered for processing a whole-slide image (WSI) in PROMETEO.} First, in the step called \textit{Read}, patches are extracted from the input WSI, and those corresponding to background are discarded (those identified as D). Then, in the next step, a score is given to each patch in order to discard patches corresponding to unwanted areas, such as pen marks and external agents. This score discriminates considering three factors: the amount of tissue that the patch contains, the percentage of pixels that are within H\&E's hue range, and the dispersion of the saturation and brightness channels. Discarded patches in this step are shown in red, while those that pass the scoring filter are highlighted in green. The third step, called \textit{Stain normalization}, applies a color normalization to the patches based on Reinhard's stain-normalization method in order to reduce color variability between samples. Finally, in the \textit{Prediction} step, each of the patches are used as input to the CNN, which classifies them as either malignant or normal tissue. A deeper insight on these steps is given in \cite{duran2020prometeo}.}
\label{fig:WholeProcess}
\end{figure*}

A set of Hematoxylin and Eosin (H\&E)-stained slides were used (158 normal \acp{WSI} and 174 malignant \acp{WSI}) provided by the Pathological Anatomy Unit of Virgen de Valme Hospital (Sevilla, Spain). These images were preprocessed using different steps. First, small subimages, called patches (100$\times$100 pixels at 10$\times$ magnification), were extracted from them. Next, background patches and patches corresponding to unwanted areas were discarded with a filter that discriminates them based on the amount of tissue that they contain, the percentage of pixels that are within H\&E's hue range, and the dispersion of the saturation and brightness channels. Then, a color normalization process called Reinhard stain-normalization \cite{reinhard2001color, magee2009colour} was applied to patches in order to reduce stain variability between samples. Finally, color-normalized patches were used as input to a \ac{CNN}, called PROMETEO, which classifies them as either malignant or normal tissue with a certain probability. A deeper insight on these steps is given in \cite{duran2020prometeo} and can be seen in Fig. \ref{fig:WholeProcess}.

Different features were obtained from PROMETEO's output in order to create the dataset. The first feature considered to discriminate between malignant or normal \ac{WSI} was the percentage of malignant tissue area, also called malignant tissue ratio (MTR), expressed between 0 and 1. This was calculated by dividing the number of patches classified as malignant by the total amount of tissue patches extracted from the \ac{WSI}. This is the most representative data to perform a slide-level classification, since the more malignant patches the network detects on the \ac{WSI}, the more likely it is of being malignant. However, based on the error of the \ac{CNN} when performing the patch classification, the percentage of malignant tissue of the \ac{WSI} should not be the only input to be considered for the patch aggregation task, since there are some exceptions that do not meet the aforementioned rule (e.g. a malignant \ac{WSI} with a small tumor in a specific region or a normal \ac{WSI} with a relatively high percentage of incorrectly-classified malignant tissue area).

\begin{figure*}[pos=htbp,width=\textwidth,align=\centering]
\centering\includegraphics[width=.85\textwidth]{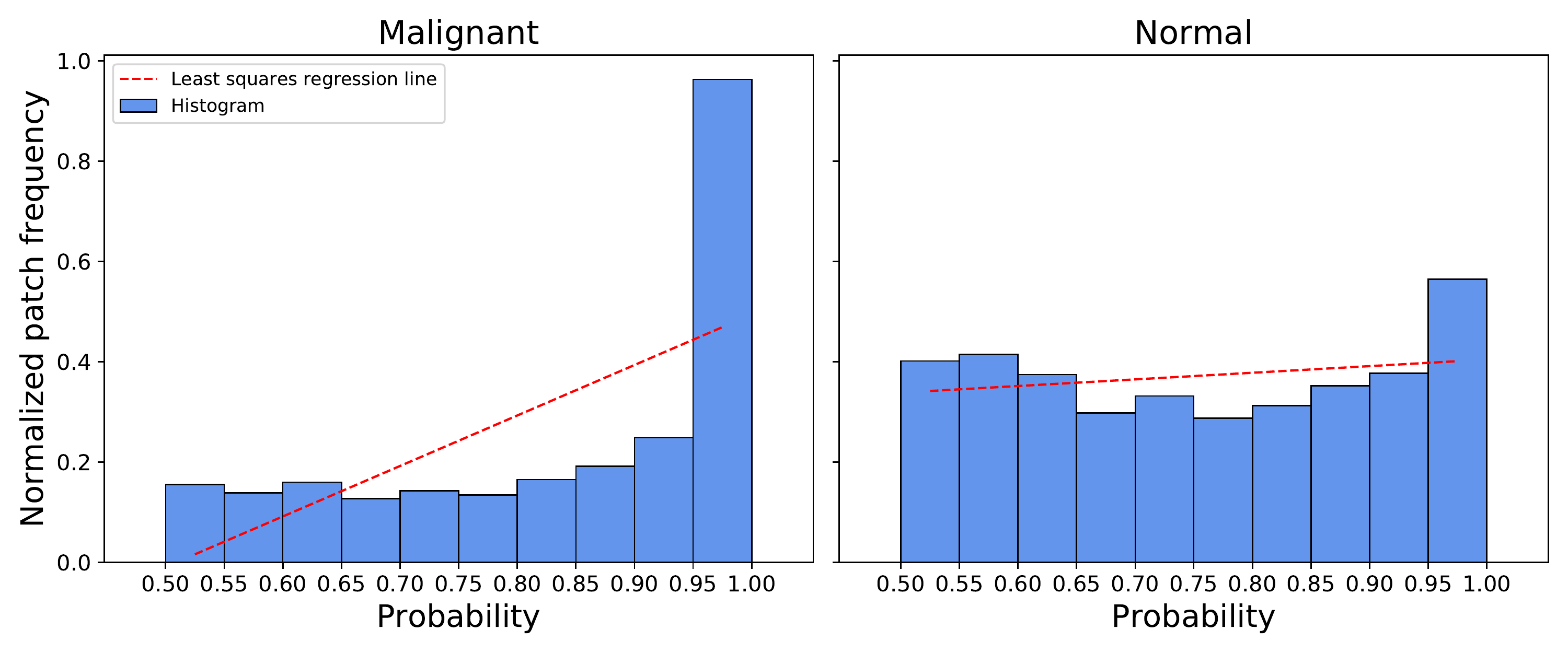}
\caption{\textbf{Mean probability histogram of the normalized patch frequency across all the WSIs, distinguishing between malignant (left) and normal (right) samples.} The least squares regression line is shown with a red dashed line. As can be seen, for malignant WSIs, the system tends to classify patches as malignant with a higher confidence. This produces a least squares regression line with a steeper slope. On the other hand, for the normal WSIs, the classification for malignant patches is not that accurate, which leads to a less steep regression line.}
\label{fig:histogram}
\end{figure*}

Therefore, another feature taken into account to distinguish between malignant and normal \acp{WSI} was the distribution of the prediction probability for malignant patches. When the \ac{CNN} predicts a patch, it reports the probability of the patch for being either malignant or normal. If we only focus on the malignant probability, the network should have a higher confidence for patches corresponding to malignant tissue than for those corresponding to normal tissue that have been incorrectly predicted as malignant. Thus, a 10-bin histogram with the prediction probabilities of the patches classified as malignant for each \ac{WSI} was calculated. These probabilities were distributed from 50\% to 100\%, with 5\% range for each bin. The histogram was normalized with respect to the total number of tissue patches. Along with the malignant probability histogram (MPH), the least squares regression line (LSRL) of the histogram, defined as $y = mx + b$, was also calculated, where m and b, which refer to the slope and the Y-intercept, are described in equations \ref{eq:leastSquaresa} and \ref{eq:leastSquaresb}, respectively. This line represents the best approximation of the set of probabilities for all malignant patches of the corresponding \ac{WSI}. The mean histogram for both malignant and normal \acp{WSI} are shown in Fig. \ref{fig:histogram} together with their corresponding LSRLs, which are highlighted in red.



\begin{equation}
\label{eq:leastSquaresa}
m = \frac{N \sum x_{i}y_{i} - \sum x_{i} \sum y_{i}}{N \sum x_{i}^{2} - (\sum x_{i})^{2}}
\end{equation}

\begin{equation}
\label{eq:leastSquaresb}
b = \frac{N \sum x_{i}^{2}\sum y_{i} - \sum x_{i} \sum x_{i}y_{i}}{N \sum x_{i}^{2} - (\sum x_{i})^{2}}
\end{equation}

Where $x$ and $y$ represent the coordinates of the different values of the histogram.


As it was previously mentioned, the error of the \ac{ML} algorithm (a \ac{CNN} in this case) leads to errors in the classification, which in a \ac{WSI} is presented as sparse normal tissue patches being classified as malignant. Therefore, in a \ac{WSI} diagnosed as normal, patches classified as malignant by the \ac{CNN} are sparsely distributed through the tissue. On the other hand, in a cancerous \ac{WSI}, malignant-classified patches tend to be focused around the tumor areas. Due to this reason, the dispersion factor of malignant-classified patches was also considered as another relevant input for the slide-level classification between normal and malignant \acp{WSI}. This factor was obtained by calculating the number of malignant connected components (MCC), which counts the isolated components (sets of malignant patches) in the classification result according to a specific distance $D$.  Algorithm \ref{alg:connected_componts_algorithm} details the method used to calculate the number of connected components based on the center coordinates of malignant patches and $D$. In this work, five different values were considered for $D$ (142, 283, 425, 566 and 708 pixels), which correspond to the Euclidean distances (i.e., radii) from a patch to a range of 1 up to 5 patches-distance, taking into account that the distance between two patches is 100 pixels (patches are 100$\times$100 pixels size). The number of connected components was normalized with respect to the total number of malignant patches for each \ac{WSI}. This way, normal samples with a low quantity of sparse misclassifications are penalized when compared to malignant samples with sparse tumoral tissue regions.

\begin{algorithm*}[]
\caption{Connected components algorithm}
\label{alg:connected_componts_algorithm}
\SetKwInOut{Input}{inputs}
\SetKwInOut{Output}{output}
\SetKwProg{ConnectedPatches}{ConnectedPatches}{}{}
\ConnectedPatches{$(centers, D)$}{
    \Input{A list of center points from malignant patches ($centers$); a distance~($D$).}
    \Output{A set of lists of connected patch centers with relative distance $D$.}
 connected\_components = []\;
 current\_component = []\;
 \While{count($centers$) > 0}{
  current\_component  = []\;
  current\_component.append($centers$[0])\;
  centers.remove(centers[0])\;
  \ForEach{center in current\_component}{
    \ForEach{point in $centers$}{
        \If{distance from center to point <= $D$}{
            current\_component.append(point)\;
            $centers$.remove(point)\;
        }
    }
  }
  connected\_components.append(current\_component)\;
 }
 }
\end{algorithm*}

\subsection{Wide \& Deep network model}
\label{subsec:wide_and_deep}


The dataset described in section~\ref{subsec:dataset} was used as input to a \ac{NN} model called W\&D \cite{cheng2016wide} to provide a slide-level classification between normal and malignant \acp{WSI}. The W\&D model combines both wide and deep components. The wide component memorizes sparse interactions between features effectively, which can be defined as learning how the output responds to combinations of sparse input values. On the other hand, the deep component corresponds to the feed-forward neural network which represents the generalization, this is, the ability to handle unseen data. Therefore, the benefits from both memorization (wide) and generalization (deep) are combined and achieved in a single model \cite{cheng2016wide}.

In this work, the malignant tissue ratio was used as the wide element while the malignant probability histogram, the slope and Y-intercept of the LSRL and the number of malignant connected components were used as the deep elements. Each of the deep data were separately connected to two hidden layers of 300 neurons. Then, these layers were concatenated together with the wide element to a hidden block of two hidden layers with 300 neurons each. Finally, this hidden block was connected to the output layer, a SoftMax function which performs the classification of the \ac{WSI} as either malignant or normal. This way, complex features are extracted from combinations of sparse inputs and then concatenated together in order to perform the final decision.

\begin{figure*}[htbp]
\centering\includegraphics[width=1\linewidth]{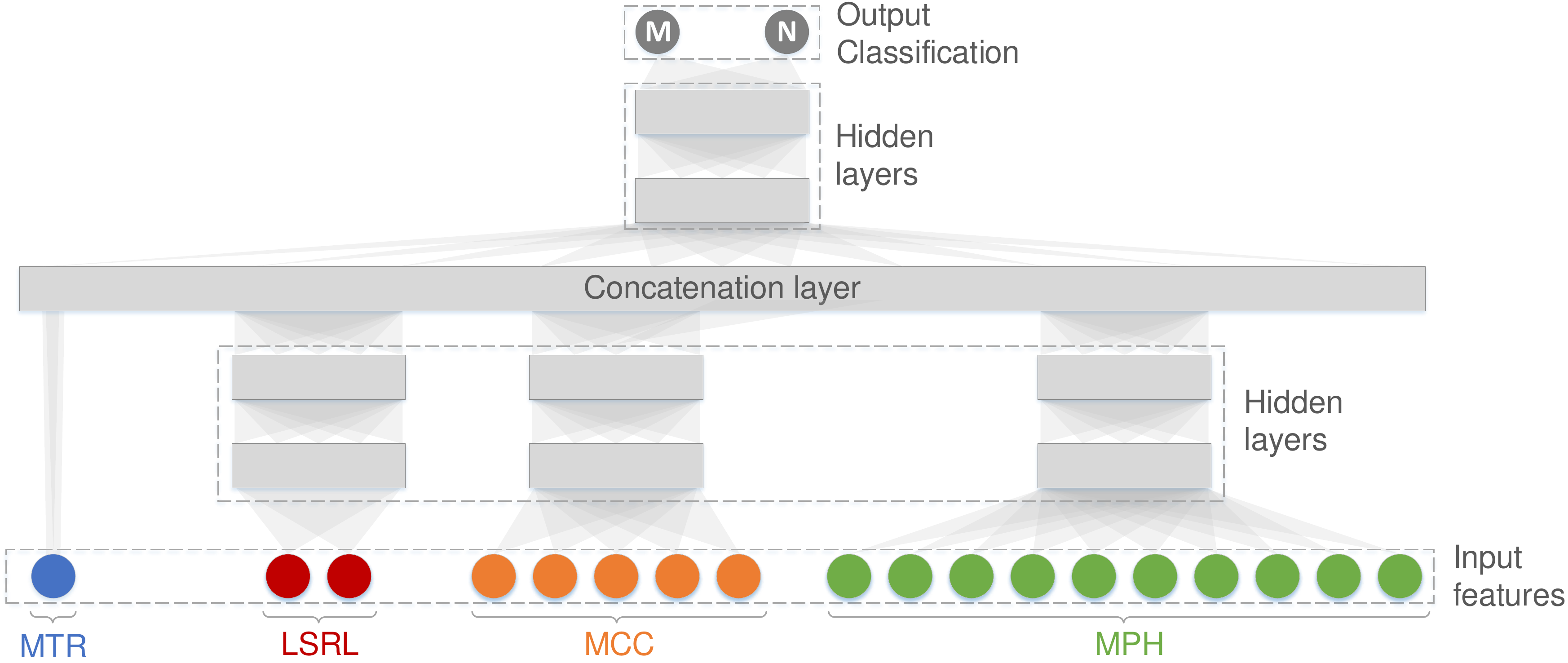}
\caption{\textbf{Diagram of the W\&D network model proposed in this work.} Each hidden layer consists of 300 neurons. The input features, which are detailed in section~\ref{subsec:dataset}, are: the malignant tissue ratio (MTR) of the WSI, the slope and Y-intercept of the least squares regression line (LSRL) of the histogram, the number of malignant connected components (MCC) with 5 different radii (from 1 to 5 malignant patch distance), and the 10-bin malignant probability histogram (MPH) between 50\% and 100\% with 5\% ticks. These input features are used to classify the WSI as either malignant (M) or normal (N).}
\label{fig:wide_and_deep}
\end{figure*}

\begin{figure*}[H]
\centering\includegraphics[width=1\linewidth]{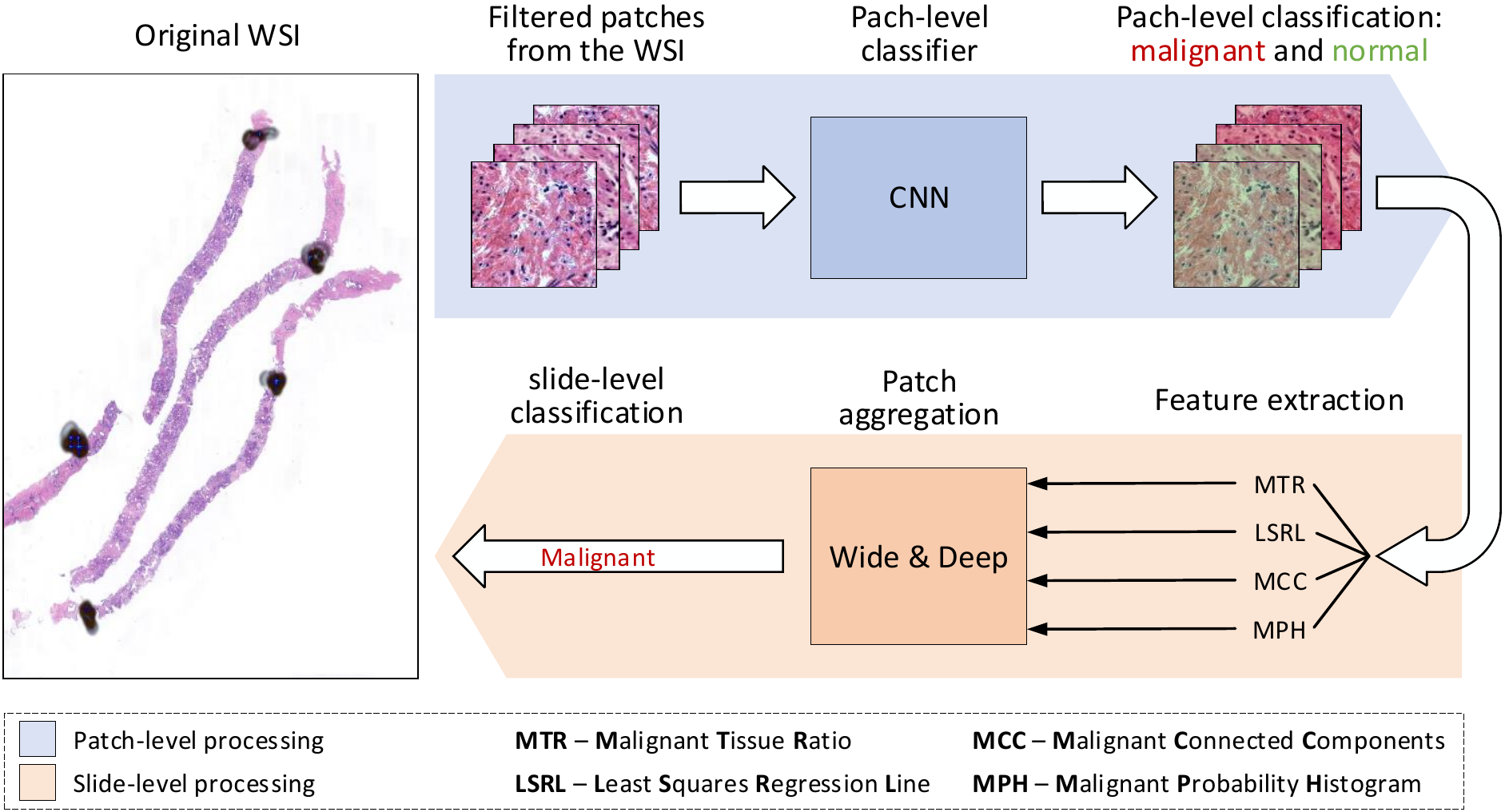}
\caption{\textbf{Diagram of the whole processing step for the PCa screening task.} First, the WSI is processed at patch level, following the same procedure presented in Figure~\ref{fig:WholeProcess}. Then, the output classification for each of the filtered patches from the original WSI is used to perform a slide-level prediction using the W\&D model presented in Figure~\ref{fig:wide_and_deep}, where the extracted features are used to classify the WSI as either malignant or normal.}
\label{fig:whole_process_CADWD}
\end{figure*}

Figure \ref{fig:wide_and_deep} depicts the custom W\&D model used in this work, where the different inputs and layers can be seen. Figure~\ref{fig:whole_process_CADWD} represents the whole processing step for the prostate screening task, highlighting both the patch-level and the slide-level components.


\subsection{Training and validation}
\label{subsec:training_and_validation}

K-fold stratified cross-validation was performed to measure the generalization ability of the model. This technique consisted in dividing the dataset in 5 sets ($K~=~5$). For each fold, the network was trained using four of the five sets (80\% of the dataset) for 10000 epochs and validated using the remaining one (20\% of the dataset). This way, for each experiment, the network was trained and validated a total of five times with different data. The final results are presented as the mean accuracy calculated over the five cross-validation folds.

To validate the network, different evaluation metrics were used. These were the accuracy (eq. \ref{eq:Accuracy}), sensitivity (eq. \ref{eq:Sensitivity}), precision (eq. \ref{eq:Precision}), F1 score (eq. \ref{eq:F1-score}) and \ac{AUC} of the \ac{ROC} curve.

\begin{equation}\label{eq:Accuracy}
    \textrm{Accuracy} = 100 \times \dfrac{\textrm{TP}+\textrm{TN}}{\textrm{TP}+\textrm{TN}+\textrm{FP}+\textrm{FN}}
\end{equation}

\begin{equation}\label{eq:Sensitivity}
    \textrm{Sensitivity} = 100 \times \dfrac{\textrm{TP}}{\textrm{TP}+\textrm{FN}}
\end{equation}

\begin{equation}\label{eq:Precision}
    \textrm{Precision} = 100 \times \dfrac{\textrm{TP}}{\textrm{TP}+\textrm{FP}}
\end{equation}

\begin{equation}\label{eq:F1-score}
    \textrm{F1 score} = 2 \times \dfrac{\textrm{Precision} \times \textrm{Sensitivity}}{\textrm{Precision} + \textrm{Sensitivity}}
\end{equation}

Where TP and FP denote true positive cases (when the model diagnoses a malignant \ac{WSI} correctly) and false positive cases (the network diagnoses a normal \ac{WSI} as malignant), respectively. TN and FN denote true negative cases (the system classifies a normal \ac{WSI} as normal) and false negative cases (the network classifies a malignant \ac{WSI} as normal), respectively.

For designing, training and validating the model, both TensorFlow\footnote{https://www.tensorflow.org  (accessed \today} and Keras\footnote{https://keras.io (accessed \today)} were used.

\section{Results and Discussion}
\label{sec:results_and_discussion}


After training the custom W\&D model (section~\ref{subsec:wide_and_deep}) with the dataset presented in section~\ref{subsec:dataset}, all the different metrics described in section~\ref{subsec:training_and_validation} were calculated and obtained in order to evaluate the proposed system. Table~\ref{tb:CV_results} summarizes the results for each cross-validation fold together with the average for all the evaluation metrics. With these, the average results were calculated, achieving an accuracy of 94.24\%, a sensitivity of 98.87\%, a precision of 90.23\%, a F1 score of 94.33\% and an \ac{AUC} of 0.94.

\begin{table}[pos=htbp,width=.48\textwidth,align=\centering]
\centering
\caption{\textbf{Validation results obtained with the proposed W\&D model.} The accuracy, sensitivity, precision, F1 score and AUC) are shown for each of the different cross-validation folds.
The average of the obtained metrics across the five folds is also presented.}
\label{tb:CV_results}
\resizebox{0.48\textwidth}{!}{%
\begin{tabular}{cccccc}
\hline
\textbf{Fold}    & \textbf{\begin{tabular}[c]{@{}c@{}}Accuracy\\ (\%)\end{tabular}} & \textbf{\begin{tabular}[c]{@{}c@{}}Sensitivity\\ (\%)\end{tabular}} & \textbf{\begin{tabular}[c]{@{}c@{}}Precision\\ (\%)\end{tabular}} & \multicolumn{1}{c}{\textbf{\begin{tabular}[c]{@{}c@{}}F1 score\\ (\%)\end{tabular}}} & \textbf{AUC}   \\ \hline
1                & 93.93                                                            & 100                                                                 & 89.74                                                             & 94.59                                                                                & 0.93          \\ \hline
2                & 93.93                                                            & 97.29                                                               & 92.30                                                             & 94.73                                                                                & 0.93          \\ \hline
3                & 95.45                                                            & 100                                                                 & 90.32                                                             & 94.91                                                                                & 0.96          \\ \hline
4                & 93.93                                                            & 100                                                                 & 87.09                                                             & 93.10                                                                                & 0.94          \\ \hline
5                & 93.93                                                            & 97.05                                                               & 91.66                                                             & 94.28                                                                                & 0.93          \\ \hline
\textbf{Average} & \textbf{94.24}                                                   & \textbf{98.87}                                                      & \textbf{90.23}                                                    & \textbf{94.33}                                                   & \textbf{0.94} \\ \hline
\end{tabular}
}

\end{table}

\begin{figure*}[pos=htbp,width=\textwidth,align=\centering]
\centering
\includegraphics[width=.77\textwidth]{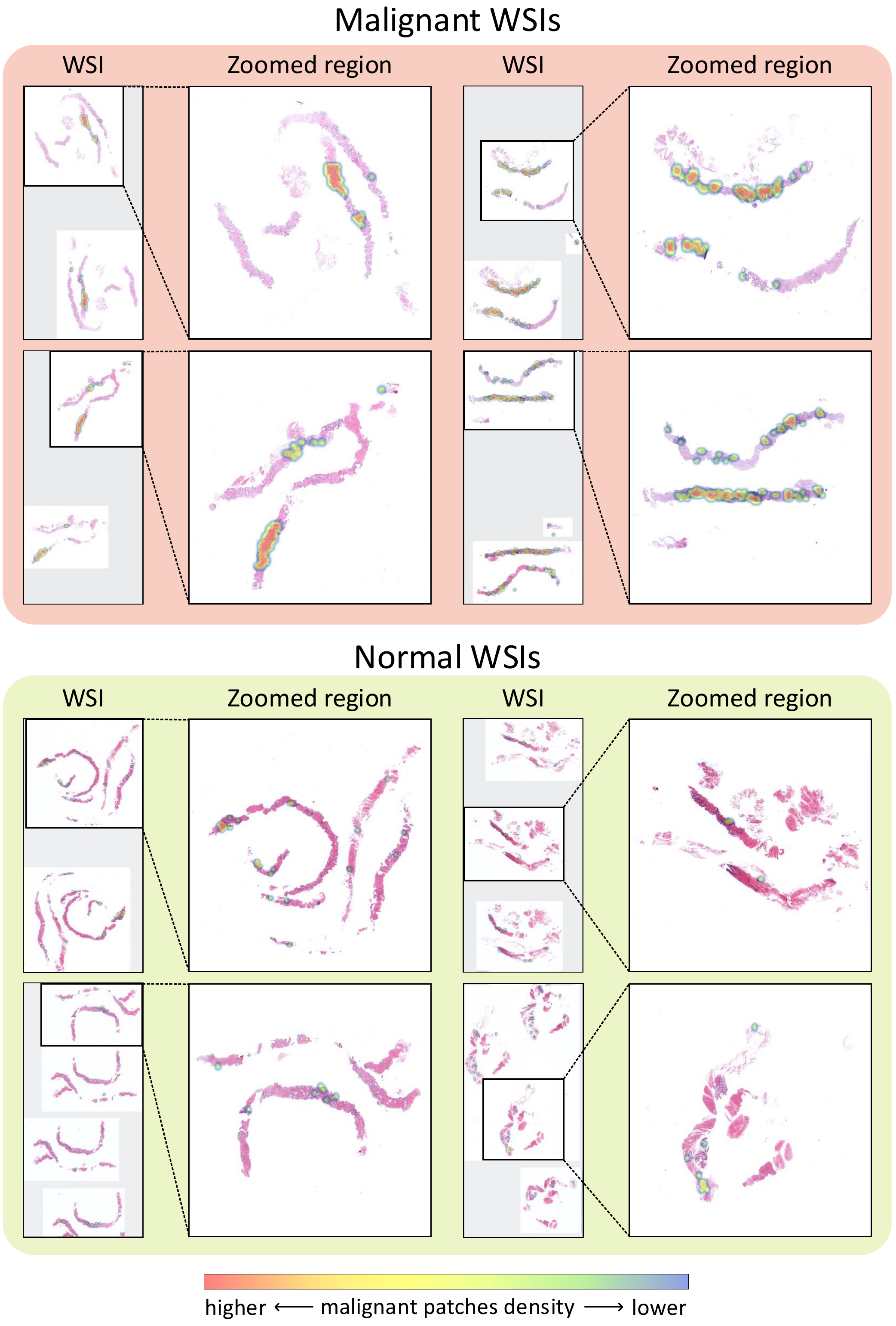}
\caption{\textbf{Eight different WSI samples extracted from the dataset presented in section~\ref{subsec:dataset}.} A heatmap of the malignant patches predicted by PROMETEO is drawn on top of the WSI, and zoomed regions are presented for better visualization. Red regions represent higher concentrations of malignant patches, while blue represent the opposite. The examples presented were correctly classified by the proposed W\&D model.}
\label{fig:normal_malignant_WSIs}
\end{figure*}

As can be seen, the results obtained across the different folds are consistent and the proposed model achieves very high scores in all the different metrics studied for this classification task, particularly in terms of sensitivity. The sensitivity, which in this field is defined as the ability of the system to identify \ac{PCa}, is of utmost importance for reporting and assessing the performance of the screening test \cite{hakama2007sensitivity}. The proposed system is able to achieve an average sensitivity of around 99\%, where three of the folds achieved perfect sensitivity (100\%). This means that our custom model makes almost no mistakes when predicting a malignant sample as such, making it a reliable patch aggregation method, together with PROMETEO, for \ac{PCa} detection in \acp{WSI}. Figure \ref{fig:normal_malignant_WSIs} shows some examples of correctly classified \acp{WSI}.

The results obtained in this study were compared with different \ac{ML}-based methods and classifiers using the same dataset. The following well-known machine learning algorithms were used to classify the \acp{WSI}: an \ac{ANN} \cite{yegnanarayana2009artificial}, a \ac{SVM} \cite{wang2005support}, a \ac{RF} \cite{breiman2001random} and a \ac{KNN} \cite{jiang2007survey}. Table~\ref{tb:results_comparison} summarizes the results obtained for each method, which are represented as the average of the evaluation metrics (see section~\ref{subsec:training_and_validation}) obtained for each cross-validation fold.


As it can be seen, the best results for accuracy, sensitivity, F1 score and \ac{AUC} are obtained with the proposed W\&D model, with the exception of precision, for which \ac{SVM} achieves the highest value. As it was previously mentioned, sensitivity is the most relevant metric for measuring the performance of a classifier when performing a screening test. In this case, the proposed architecture is the one achieving the highest sensitivity score among the different algorithms evaluated, with a difference of more than 6\% with the second highest, which is the \ac{ANN}. On the other hand, \ac{SVM} achieves around 99\% precision, which could be very relevant for other binary or multi-class classification tasks, but not as much as the sensitivity when developing a \ac{ML}-based \ac{PCa} screening method that could help experts in order to accelerate the whole process.

\begin{table*}[pos=htbp,width=1\textwidth,align=\centering]
\centering
\caption{\textbf{Validation results calculated from the average of the evaluation metrics (accuracy, sensitivity, precision, F1 score and AUC) for the 5 different cross-validation sets.} The results obtained with the proposed W\&D model are compared to other state-of-the-art ML-based algorithms, namely, ANN, SVM, RF and KNN. The best result for each specific evaluation metric is highlighted in bold.}
\label{tb:results_comparison}
\resizebox{0.7\textwidth}{!}{%
\begin{tabular}{cccccc}
\hline
\textbf{Model} & \textbf{\begin{tabular}[c]{@{}c@{}}Accuracy\\ (\%)\end{tabular}} & \textbf{\begin{tabular}[c]{@{}c@{}}Sensitivity\\ (\%)\end{tabular}} & \textbf{\begin{tabular}[c]{@{}c@{}}Precision\\ (\%)\end{tabular}} & \textbf{\begin{tabular}[c]{@{}c@{}}F1 score\\ (\%)\end{tabular}} & \textbf{AUC}   \\ \hline
W\&D (proposed)          & \textbf{94.24}                                                   & \textbf{98.87}                                                      & 90.23                                                             & \textbf{94.33}                                                   & \textbf{0.94} \\ \hline
ANN            & 89.69                                                            & 92.47                                                               & 87.29                                                             & 89.54                                                            & 0.89          \\ \hline
SVM            & 88.18                                                            & 80.78                                                               & \textbf{98.76}                                                    & 88.79                                                            & 0.89          \\ \hline
RF             & 88.84                                                            & 84.89                                                               & 92.23                                                             & 88.22                                                            & 0.88          \\ \hline
KNN            & 88.48                                                            & 83.29                                                               & 94.31                                                             & 88.31                                                            & 0.88          \\ \hline
\end{tabular}
}

\end{table*}


    

\section{Conclusions}
\label{sec:conclusions}

In this work, the authors present a novel \ac{ML}-based method to classify \acp{WSI} of prostate tissue as normal or malignant at global slide level based on a previous patch-level classification. This classification is based on a novel \ac{NN} model called W\&D, which combines both linear model components (wide) and neural network components (deep) in order to achieve both memorization and generalization in a single model. The custom W\&D proposed model  classifies each \ac{WSI} as normal or malignant considering different processed inputs. This information was obtained using PROMETEO, a \ac{CAD} system which extracts small patches from \acp{WSI} which are first preproccessed and then classified, reporting a heatmap that shows where malignant areas are located inside the corresponding \ac{WSI}. From the information obtained from malignant patches, different processed features are calculated, which are then used as input to the proposed W\&D model. These are the malignant tissue ratio, the 10-bin malignant probability histogram between 50\% and 100\% with 5\% ticks, the slope and Y-intercept of the least squares regression line of the histogram and the number of malignant connected components with 5 different radii. The network was trained and validated using 5-fold cross-validation. The average results obtained for the cross-validation sets with the W\&D model achieved an accuracy of 94.24\%, a sensitivity of 98.87\%, a precision of 90.23\%, a F1 score of 94.33\% and an \ac{AUC} of 0.94. The proposed model was compared with other state-of-the-art methods (\ac{ANN}, \ac{SVM}, \ac{RF} and \ac{KNN}) using the same dataset. The results show that the W\&D model performs better in terms of accuracy, sensitivity, F1 score and \ac{AUC}. The promising results obtained with this novel model show that the proposed system could aid pathologists when analyzing histopathological images as a screening method, discriminating between normal and malignant \ac{PCa} slides.

\section*{Acknowledgments}

We would like to thank Antonio Felix Conde-Martin and the Pathological Anatomy Unit of Virgen de Valme Hospital in Seville (Spain) for their support in the PROMETEO project, together with VITRO S.A., along with providing us with annotated \acp{WSI} from the same hospital. The authors would like to thank the Spanish Agencia Estatal de Investigación (AEI) for supporting this work. This work was partially funded by Spanish Agencia Estatal de Investigación (AEI) project MINDROB (PID2019- 105556GB-C33/ AEI/10.13039/501100011033), with support from the European Regional Development Fund, by the EU H2020 project CHIST-ERA SMALL (PCI2019-111841-2) and~by the Andalusian Regional Project PAIDI2020 (with FEDER support) PROMETEO (AT17\_5410\_USE).






\bibliographystyle{unsrt}
\bibliography{cas-refs}

\end{document}